\documentclass[letterpaper]{article} % DO NOT CHANGE THIS
\usepackage{aaai2026}  % DO NOT CHANGE THIS
\usepackage{times}  % DO NOT CHANGE THIS
\usepackage{helvet}  % DO NOT CHANGE THIS
\usepackage{courier}  % DO NOT CHANGE THIS
\usepackage[hyphens]{url}  % DO NOT CHANGE THIS
\usepackage{graphicx} % DO NOT CHANGE THIS
\usepackage{natbib}  % DO NOT CHANGE THIS AND DO NOT ADD ANY OPTIONS TO IT
\usepackage{caption} % DO NOT CHANGE THIS AND DO NOT ADD ANY OPTIONS TO IT
\usepackage{amssymb} % For \checkmark
\usepackage{algorithm}
\usepackage{algorithmic}
\usepackage{booktabs}
\usepackage{multirow}
\usepackage{graphicx}
\usepackage{makecell}
\usepackage{xcolor,colortbl}
\usepackage{subfigure}
\usepackage{mathrsfs}
\usepackage{amsmath}
\usepackage{amssymb}
\usepackage{pifont}

\definecolor{uncolor}{RGB}{222,235,247}  % 浅蓝
\definecolor{fullcolor}{RGB}{232,245,233}  % 浅绿
\definecolor{weakcolor}{RGB}{255,248,220}  % 浅棕
\definecolor{light-gray}{gray}{0.9}

\makeatletter\def\@listi{\leftmargin\leftmargini \topsep .5em \parsep .5em \itemsep .5em}
\def\@listii{\leftmargin\leftmarginii \labelwidth\leftmarginii \advance\labelwidth-\labelsep \topsep .4em \parsep .4em \itemsep .4em}
\def\@listiii{\leftmargin\leftmarginiii \labelwidth\leftmarginiii \advance\labelwidth-\labelsep \topsep .4em \parsep .4em \itemsep .4em}\makeatother

\newcounter{checksubsection}
\newcounter{checkitem}[checksubsection]

\usepackage{newfloat}
\usepackage{listings}
\DeclareCaptionStyle{ruled}{labelfont=normalfont,labelsep=colon,strut=off} % DO NOT CHANGE THIS
\floatstyle{ruled}
\newfloat{listing}{tb}{lst}{}
\floatname{listing}{Listing}
\title{Weakly-Supervised Image Forgery Localization via Vision-Language Collaborative Reasoning Framework}

\author {
    Ziqi Sheng\textsuperscript{\rm 1},
    Junyan Wu\textsuperscript{\rm 1},
    Wei Lu\textsuperscript{\rm 1}\thanks{Corresponding Author},
    Jiantao Zhou\textsuperscript{\rm 2}
}
\affiliations {
    \textsuperscript{\rm 1}School of Computer Science and Engineering, MoE Key Laboratory of Information Technology,\\
Guangdong Province Key Laboratory of Information Security Technology, Sun Yat-sen University.\\
    \textsuperscript{\rm 2}State Key Laboratory of Internet of Things for Smart City, \\
    Department of Computer and Information Science, University of Macau\\
    shengzq@mail2.sysu.edu.cn, wujy298@mail2.sysu.edu.cn, luwei3@mail.sysu.edu.cn, jtzhou@um.edu.mo
}

\begin{document}

\maketitle

\begin{abstract}

Image forgery localization aims to precisely identify tampered regions within images, but it commonly depends on costly pixel-level annotations. 
To alleviate this annotation burden, weakly supervised image forgery localization (WSIFL) has emerged, yet existing methods still achieve limited localization performance as they mainly exploit intra-image consistency clues and lack external semantic guidance to compensate for weak supervision.
In this paper, we propose ViLaCo, a vision-language collaborative reasoning framework that introduces auxiliary semantic supervision distilled from pre-trained vision-language models (VLMs), enabling accurate pixel-level localization using only image-level labels. 
Specifically, ViLaCo first incorporates semantic knowledge through a vision-language feature modeling network, which jointly extracts textual and visual priors using pre-trained VLMs. 
Next, an adaptive vision-language reasoning network aligns textual semantics and visual features through mutual interactions, producing semantically aligned representations.
Subsequently, these representations are passed into dual prediction heads, where the coarse head performs image-level classification and the fine head generates pixel-level localization masks, thereby bridging the gap between weak supervision and fine-grained localization.
Moreover, a contrastive patch consistency module is introduced to cluster tampered features while separating authentic ones, facilitating more reliable forgery discrimination.
Extensive experiments on multiple public datasets demonstrate that ViLaCo substantially outperforms existing WSIFL methods, achieving state-of-the-art performance in both detection and localization accuracy.

\end{abstract}

\section{Introduction}

\begin{figure}[]
    \centering
    \subfigure[Fully-supervised training.]{%
        \includegraphics[width=0.48\textwidth]{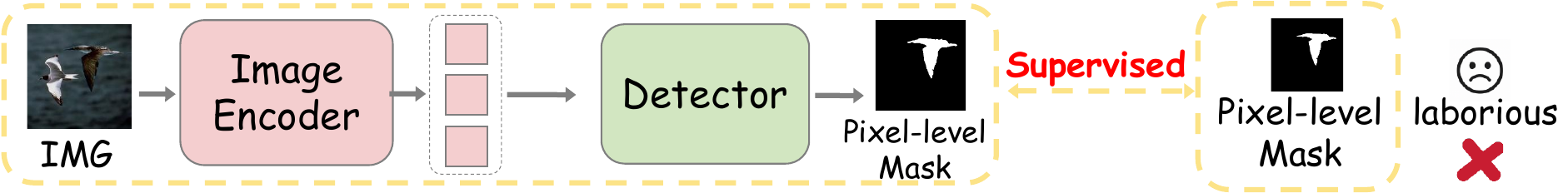}%
        \label{fig: supervised}
    }
    \subfigure[Weakly-supervised training.]{%
        \includegraphics[width=0.48\textwidth]{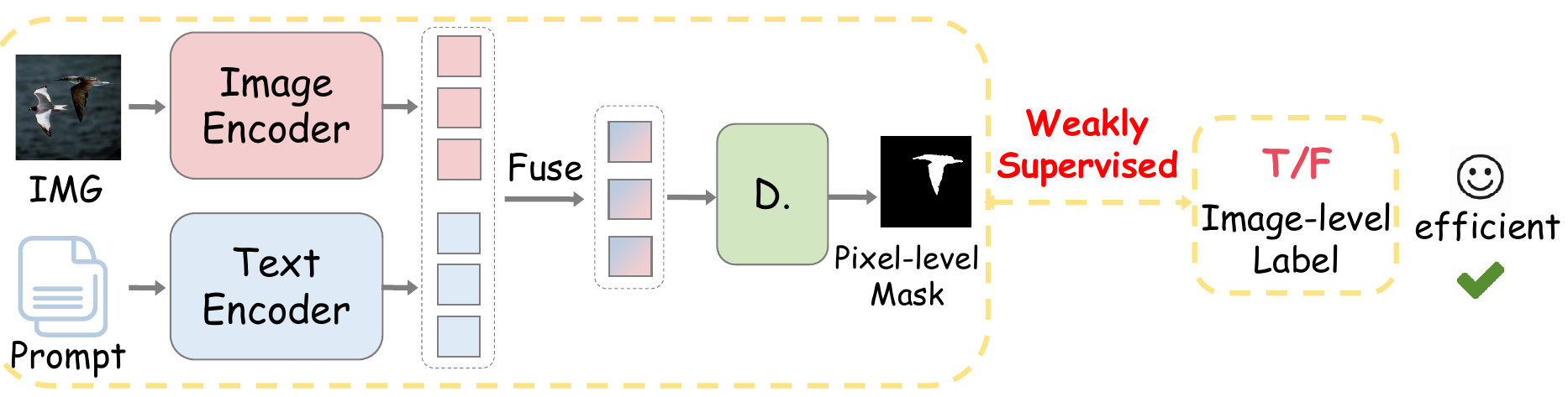}%
        \label{fig: weakly supervised}
    }
    \caption{Difference between fully-supervised and weakly-supervised training strategies.
(a) Fully-supervised methods use pixel-level masks for both training and prediction.
(b) Weakly-supervised methods are trained only with binary image-level labels but are still required to predict pixel-level manipulation masks. Besides, the proposed method further leverages text prompts for extra supervision.}
    \label{fig:different_trainning}
\end{figure}

With the rapid development of AI-generated content technologies, image forgery has become increasingly common, and the era of “what you see is what you get” is gradually fading.
The proliferation of malicious forged images is undermining public trust and posing serious threats to economic security and public safety.
This drives the urgent need for reliable forgery localization methods that can accurately identify tampered regions.
Although numerous approaches have been explored~\cite{lou2025exploring, sheng2024dirloc}, most methods rely on fully supervised learning, requiring costly pixel-level annotations during training (Fig. \ref{fig: supervised}).
However, due to the rapid growth of forged images, collecting large-scale, high-quality pixel annotations is labor-intensive and costly, making it hard to apply in real-world settings. 

To tackle these challenges, weakly supervised image forgery localization (WSIFL) has emerged as a promising direction. 
WSIFL methods aim to localize pixel-level manipulations using only image-level binary labels, typically leveraging contrastive learning and self-supervision to discover tamper-sensitive regions and enable fine-grained prediction.
For instance, \cite{zhai2023towards} utilizes multi-scale consistency and inter-block correlations to discover the manipulation regions. 
\cite{li2025m2rl} employs block-level self-consistency and frame-level contrastive learning to distinguish consistent and inconsistent regions within images.
Although these methods have achieved certain progress, they rely solely on internal image signals and lack external semantic guidance, which limits their ability to accurately identify forged regions.

Meanwhile, vision-language models \cite{radford2021learning} have demonstrated remarkable capacity in aligning visual and textual modalities and improving downstream tasks like image classification and retrieval.
However, in the field of WSIFL, its potential remains underexplored.
In this paper, we aim to leverage the rich semantic alignment relationships between visual and textual modalities provided by pre-trained vision-language models (VLMs) to offer extra guidance for the WSIFL task (Fig. \ref{fig: weakly supervised}).
Although VLMs provide a powerful external knowledge source, directly adapting them to the WSIFL task still faces three important challenges. 
(1) Modeling forgery-aware representations: how to effectively learn fine-grained, forgery-aware features without explicit pixel-level supervision.
(2) Fusing cross-modal features: how to enable effective interactions between textual semantics and visual clues to accurately highlight manipulated regions under weak supervision.
(3) Bridge supervision granularity gap: how to address the discrepancy between coarse-grained image-level labels and the required pixel-level masks.

% (1) Insufficient supervision. WSIFL only provides coarse image-level labels, and relying solely on internal visual cues often fails to reveal subtle manipulation traces. 
% To address this, we introduce auxiliary textual prompts derived from binary labels to provide additional semantic supervision.
% (2) Effective semantic guidance. Even with textual supervision, how to make text features adaptively guide the model in mining image-internal anomalies remains non-trivial. 
% We thus design an adaptive vision-language reasoning module that enables fine-grained interaction between textual semantics and visual features.
% (3) Supervision granularity gap. There is a mismatch in granularity between image-level labels and the required pixel-level localization masks.
% To alleviate this gap, we adopt a coarse-to-fine dual prediction strategy, combining patch-level anomaly aggregation for image classification and a mask decoder for precise manipulation localization.

To overcome these challenges, we propose ViLaCo, a vision-language collaborative reasoning framework that progressively injects semantic knowledge into the localization process. 
Specifically, to address the first challenge, we introduce the vision-language feature modeling network to jointly model a forgery-aware prompt and image representations. In addition, a lightweight local-global spatial adapter is used to capture image forgery traces by modeling both local inconsistencies and global dependencies.
To address the second challenge, we design the adaptive
vision-language reasoning network, containing an attention-based reasoning layer and a forgery-aware feature aggregator to strengthen the interaction between textual and visual features. 
Finally, to address the third challenge,  we adopt a dual prediction strategy, including a coarse branch aggregating patch-level scores for image-level anomaly detection and a fine branch combining a text-image similarity map with a mask decoder to predict pixel-level manipulation masks. 
Furthermore, a contrastive patch consistency loss is incorporated to improve the discriminability of forgery features and enhance localization accuracy.
Extensive experimental results over five public testing datasets demonstrate that ViLaCo significantly outperforms the state-of-the-art competing algorithm.

In summary, our contributions are as follows:

(1) We propose ViLaCo, a novel vision-language collaborative reasoning framework for WSIFL. 
By introducing vision-language semantic supervision and a dual-branch architecture, ViLaCo simultaneously achieves effective image-level classification and accurate pixel-level localization without pixel-level annotations.

% (2) We design two key modules to fully exploit vision-language semantic guidance: the vision-language feature modeling stage that jointly encodes images and label-derived text prompts with learnable tokens while capturing manipulation clues via a lightweight local-global spatial consistency adapter, and an adaptive cross-modal reasoning module that aligns and fuses text and image features to highlight forged regions under weak supervision.

% (2) We design two key components to leverage vision-language semantic guidance: the vision-language feature modeling that jointly encodes images and text prompts with learnable tokens, and a local-global spatial consistency adapter that captures manipulation clues. The adaptive cross-modal reasoning module then aligns and fuses text and image features to highlight forged regions under weak supervision.

(2) We design a vision-language feature modeling and an adaptive vision-language reasoning to fully leverage the semantic guidance. 
The former models images and text features through a local-global spatial consistency adapter (LGS adapter) and learnable prompts, respectively. 
Subsequently, the latter aligns and fuses text and image features under weak supervision to highlight the forged areas.

(3) We introduce a contrastive patch consistency constraint that clusters tampered patch features and separates authentic ones, enhancing spatial coherence and improving the quality of predicted masks.

\section{Related Works}

\subsection{Image Manipulation Localization}

Most existing image manipulation localization techniques have been developed under fully supervised settings, aiming to detect and localize forged content at both image and pixel levels.
Early methods mainly relied on handcrafted forensic clues, which proved inadequate for handling the diverse and complex nature of real-world manipulations.
To overcome these limitations, recent studies have shifted toward more general tampering detection \cite{sheng2025sumi, triaridis2024exploring}, exploiting multiple forensic traces such as JPEG artifacts, edge inconsistencies, noise patterns, and camera fingerprints.
To better capture subtle manipulation artifacts, researchers have explored non-RGB feature domains. Noise residual views generated by SRM or learnable filters \cite{li2024unionformer, zeng2024mgqformer, guo2023hierarchical} and frequency information via DCT coefficients \cite{kwon2022learning, wang2022objectformer} have been incorporated to highlight low-level inconsistencies.
More recently, contrastive learning techniques \cite{lou2025exploring, niloy2023cfl} have been introduced to learn discriminative embeddings for authentic and forged regions.
Although these fully supervised methods are effective, they are highly dependent on pixel-level annotations, which makes training laborious and costly.
To reduce annotation costs, weakly supervised methods \cite{li2025m2rl, zhai2023towards, zhou2024exploring} aim to infer fine-grained localization from image-level labels.
However, they mainly utilize supervisory signals within the image, which limits their pixel-level localization capabilities.
In contrast, our method utilizes semantic supervision distilled from pre-trained vision-language models, introducing semantic guidance beyond internal image clues to better uncover forgery traces.

\subsection{Vision-Language Pre-training}

Vision-language pre-training has shown strong ability to align visual and textual semantics by learning from large-scale image-text pairs. 
CLIP, as a representative model, demonstrates impressive generalization across classification, detection, captioning, and dense prediction tasks \cite{zhou2022learning, zhou2022detecting, barraco2022unreasonable, rao2022denseclip}.
Recent studies further adapt these pre-trained models to specialized domains, including audio temporal forgery localization \cite{wu2025weakly}, video understanding \cite{luo2022clip4clip}, semantic segmentation \cite{kweon2024sam} and so on.
Inspired by these advances, we explore leveraging CLIP’s rich vision-language representations for weakly supervised image forgery localization. 
Instead of relying purely on image-internal clues, our method brings in semantic supervision from pre-trained vision-language models, making weakly supervised forgery localization more precise and semantically informed.

\begin{figure*}
\centering
\includegraphics[width=0.95\textwidth]{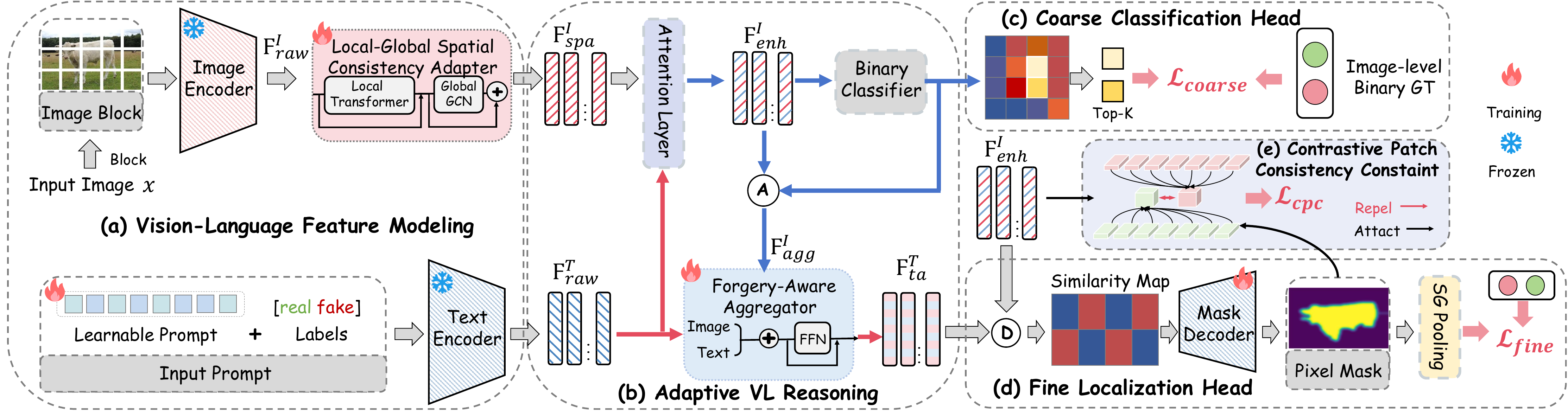}
\caption{The proposed ViLaCo framework consists of (a) vision-language feature modeling, (b) adaptive vision-language reasoning, (c) coarse classification head, (d) fine localization head, and (e) a contrastive patch consistency constraint.}
\label{fig:framework}
\end{figure*}

\section{Method}

\subsection{Problem Definition}

The weakly-supervised image forgery localization (WSIFL) task supposes that only image-level labels are available during the training stage, and encourages the models to predict whether each pixel is tampered at the inference stage.
Mathematically, given a set of samples {$\mathcal{X}$, $\mathcal{Y}$, $\mathcal{M}$},
where $\mathcal{X}$, $\mathcal{Y}$, $\mathcal{M}$ denote the sets of image, image-level binary label, and pixel-level localization mask, respectively.
For each input image $x \in \mathcal{R}^{H \times W \times C}$, it has two corresponding labels, namely, $y$ and $m$, with $m$ only for the inference stage.
Here $y \in \{0,1\}$ and $y =1$ indicates that $x$ is a tampered image; and $m \in \{0,1\}^{H \times W \times 1}$, $m = 1$ indicated the tampered regions.

\subsection{Overview}

As illustrated in Fig.~\ref{fig:framework}, we propose ViLaCo, the vision-language collaborative reasoning framework for weakly supervised image forgery localization (WSIFL). 
Unlike previous weakly supervised methods that relied solely on internal image signals, ViLaCo exploits the powerful semantic alignment between visual and textual modalities from a pre-trained vision-language model, enabling precise pixel-level localization under weak supervision. 
The framework consists of four key components:

(i) Vision-Language Feature modeling. The input image is partitioned into blocks and passed through a frozen image encoder to extract raw visual features $\mathcal{F}^I_{raw}$, which are subsequently refined by a learnable local-global spatial adapter (LGS-adapter) to capture spatial forgery clues, yielding $\mathcal{F}^I_{spa}$. For textual input, a learnable prompt combined with binary class labels is processed by a frozen text encoder to generate text features $\mathcal{F}^T_{raw}$.

(ii) Adaptive Vision-Language Reasoning. To fully leverage cross-modal information, ViLaCo adaptively fuses $\mathcal{F}^T_{raw}$ and $\mathcal{F}^I_{spa}$, producing text-enhanced visual features $\mathcal{F}^I_{enh}$ via a attention layer. 
Meanwhile, a forgery-aware aggregator jointly models visual-text interactions to obtain a tampering-aware textual embedding $\mathcal{F}^T_{ta}$. This adaptive reasoning facilitates discriminative feature learning for subsequent predictions.

(iii) Dual-Branch Coarse-to-Fine Architecture. To bridge the gap between image-level supervision and pixel-level localization, ViLaCo introduces a dual prediction head design. The coarse classification head estimates the forgery likelihood by aggregating top-$K$ suspicious patches, enabling robust binary classification. Subsequently, the fine localization head constructs a similarity map between $\mathcal{F}^I_{enh}$ and $\mathcal{F}^T_{enh}$, which is decoded into a pixel-level mask via a mask decoder.
Besides, a soft-gated pooling layer further converts the mask into an auxiliary binary prediction, thereby enabling supervision using binary labels.

(iv) Contrastive Patch Consistency Constraint. To refine localization without pixel-level ground truth, A novel contrastive consistency constraint $\mathcal{L}_{cpc}$ is proposed to ensure that patches with similar forged clues in $\mathcal{F}^T_{enh}$ are pulled closer together, while patches with significant differences are pushed apart. This mechanism encourages features to cluster around manipulated regions, thereby improving mask quality.

% Through the collaboration of visual-language modeling, reasoning, and a dual-branch design, ViLaCo achieves reliable forgery detection and fine-grained localization solely from image-level labels.

Through its vision-language modeling, adaptive reasoning, and dual-branch architecture, ViLaCo effectively bridges the gap between weak supervision and fine-grained localization, achieving accurate manipulation masks and reliable forgery detection without requiring pixel-level annotations.

\subsection{Vision-Language Feature modeling}

The vision-language feature modeling phase aims to provide reliable and modality-specific representations of both image and text inputs, serving as the foundation for subsequent cross-modal reasoning and localization. 
% Unlike other WSIFL methods, which relies solely on visual information, we introduce a parallel modeling design that independently extracts visual and textual features, ensuring that each modality preserves its intrinsic characteristics before fusion.
For the visual modality, a frozen image encoder is employed to obtain raw image embeddings $\mathcal{F}^{I}_{raw}$, which are further refined through local-global spatial adapter (LGS-adapter). 
% This adapter enhances spatial representation by leveraging a local transformer to capture fine-grained regional inconsistencies and a global graph convolution network (GCN) to model long-range structural relations, yielding spatially enriched features $\mathcal{F}^{I}_{spa}$.  
Meanwhile, for the textual modality, we directly concatenate trainable prompt embeddings with the fixed label tokens (“real” and “fake”), forming adaptive text representations.
% These augmented prompts are encoded using a frozen text encoder to produce adaptable textual embeddings $\mathcal{F}^{T}_{raw}$, allowing the text representation to better adapt to varying image content during training.  

\subsubsection{Local-Global Spatial Consistency Adapter}

To effectively encode spatial forgery clues, we propose an LGS-adapter to model both local inconsistencies and global structural dependencies within the encoded image feature. 
LGS-adapter sequentially integrates a local transformer encoder and a lightweight graph convolutional network (GCN) \cite{chen2020simple}, enabling spatially locally and globally aware feature learning.

Given the patch-level visual features $\mathcal{F}^I_{raw} \in \mathbb{R}^{n \times d}$ extracted from a frozen image encoder, where $n$ denotes the number of image patches and $d$ is the feature dimension, we first capture local spatial interactions $X_l$ using a windowed transformer encoder. 
Unlike standard transformers that compute self-attention globally, this layer restricts attention computation to non-overlapping and partially overlapping spatial windows, focusing on nearby patch interactions without cross-window message passing. 
This local design simulates convolutional receptive fields, enhancing the network's sensitivity to subtle manipulations while significantly reducing computational cost. 

While the local transformer is effective for patch-level forgery clues, it lacks the ability to model long-range spatial dependencies that may arise in complex tampering scenarios. 
To complement this, we incorporate a global reasoning step via a lightweight GCN module. 
We construct two adjacency matrices, $H_{sim}$ for capturing pairwise patch similarities and $H_{dis}$ for modeling relative spatial distances between patches.
Both matrices are row-normalized using softmax to ensure stable feature aggregation. 
The GCN propagates information across all patches to produce globally consistent features:
\begin{equation}
    X_g = \text{GELU}\big( [\text{Softmax}(H_{sim}); \text{Softmax}(H_{dis})] \, X_l W \big),
\end{equation}
where $W$ is a learnable transformation matrix. 
This global aggregation enriches each patch feature with context from distant regions, allowing the model to capture spatially coherent manipulation patterns even when tampered areas are spatially separated. 
By adaptively combining local attention and global graph reasoning, the LGS-adapter produces forgery-aware features $\mathcal{F}^I_{spa}$ that are both locally discriminative and globally consistent, providing a strong foundation for subsequent weakly supervised localization.

\subsubsection{Learnable Prompt}

In weakly supervised image forgery localization, the textual labels ( "real" and "fake") are insufficient to fully express the complex semantic information associated with image manipulations.
Such minimal labels can limit the transferability of textual embeddings and hinder effective vision-language interaction. 
Inspired by recent advances in prompt learning \cite{zhou2022learning}, we enhance the textual representation by introducing a set of learnable context tokens, forming a more adaptable and forgery-aware prompt.

Specifically, the discrete binary label is first tokenized using a pre-trained CLIP tokenizer to obtain an initial class token $t_{init} = \text{Tokenizer}(\text{label})$, where \text{label} $\in \{\text{real}, \text{fake}\}$. 
To construct a forgery-aware prompt, we concatenate $t_{init}$ with a sequence of learnable context tokens $\{c_1, \ldots, c_l\}$, yielding:
\begin{equation}
    t_p = \{c_1, \ldots, t_{init}, \ldots, c_l\}.
\end{equation}

The class token is placed at the center of the sequence to promote balanced contextualization from both directions. 
The combined prompt is then fused with positional embeddings to preserve token order and spatial relevance. 
Finally, the text encoder of CLIP processes $t_p$ and produces the enhanced textual embedding $t_{out} \in \mathbb{R}^d$. 
By adaptively learning the forgery-aware prompt $t_p$, this module enables the textual representation to better align with diverse manipulation patterns, thus facilitating effective cross-modal reasoning and improving the performance of WSIFL.

\subsection{Adaptive Vision-Language Reasoning}

\begin{figure}
\centering
\includegraphics[width=0.32\textwidth]{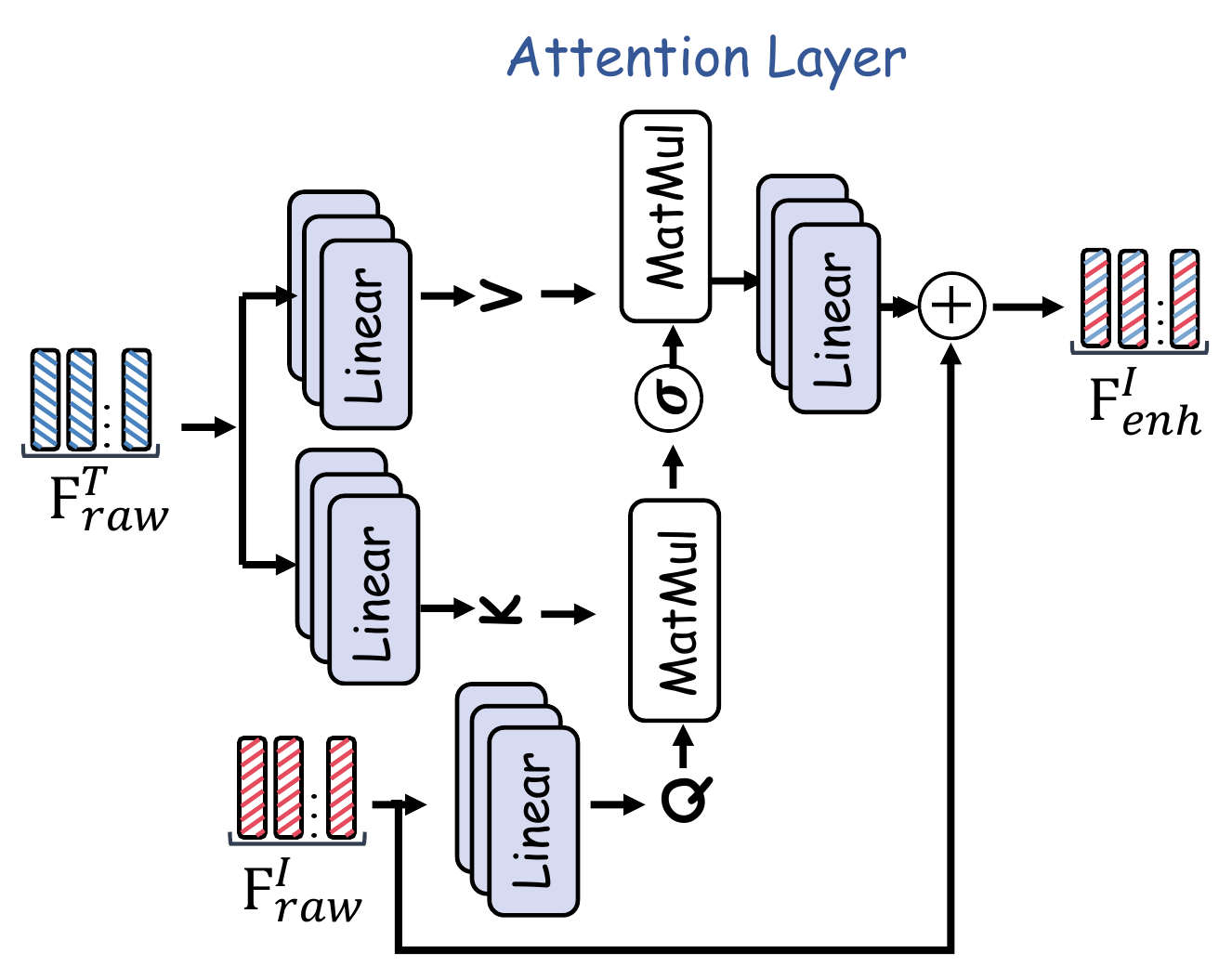}
\caption{Illustration of the structure attention layer, where textual features $F^T_{raw}$ guide the enhancement of visual features $F^I_{raw}$.}
\label{fig:attentionlayer}
\end{figure}

To effectively bridge the semantic gap between visual and textual modalities under weak supervision, we design an adaptive vision-language reasoning module. 
This module serves two purposes: (i) adaptively enhancing visual features with semantic clues from textual embeddings, and (ii) integrating cross-modal information to construct a similarity map for fine-grained forgery localization. 
% Unlike conventional multimodal fusion methods that simply concatenate features, our approach leverages text-guided cross-attention to highlight tampered regions and subsequently aligns features through contrastive reasoning. 
% This design enables robust pixel-level localization even when only image-level labels are available.

As illustrated in Fig.~\ref{fig:attentionlayer}, given the raw textual embedding $F^T_{raw}$ and the patch-level visual features $F^I_{raw}$ obtained from the LGS-adapter, we employ a text-guided attention layer to selectively enhance visual representations. 
Concretely, $F^I_{raw}$ serves as a query to attend to $F^T_{raw}$, allowing the model to amplify patches that are semantically related to tampering clues while suppressing irrelevant regions to obtain the text-enhanced visual features $F^I_{enh}$. 
This step enriches visual representations with manipulation-aware semantics, thereby enhancing the discriminative nature of feature $F^I_{enh}$.

To further strengthen cross-modal alignment, we introduce a forgery-aware aggregator that refines textual embeddings based on aggregated visual context. 
Specifically, we first summarize the enhanced visual features $F^{I}_{enh}$ into a compact visual context vector $F^{I}_{agg}$ via a soft attention pooling mechanism that highlights the most manipulation-indicative patches. 
This aggregated feature $F^{I}_{agg}$ is then fused with the raw textual embedding through a feed-forward network (FFN):
\begin{equation}
    F^T_{ta} = \text{FFN}(F^{I}_{agg} + F^T_{raw}) + F^{I}_{agg} + F^T_{raw},
\end{equation}
where $F^T_{ta}$ denotes the tampering-aware textual embedding. 
The forgery-aware aggregator enables bidirectional interaction between modalities, allowing textual features to dynamically adapt to image-specific manipulation patterns. 
This tampering-aware textual embedding $F^T_{ta}$, together with $F^I_{enh}$, provides strong discriminative signals for downstream coarse classification and fine-grained localization.

\subsection{Dual-Branch Coarse-to-Fine Architecture}

To bridge the gap between image-level supervision and pixel-level tampering localization, we design a dual-branch coarse-to-fine architecture. 
This component decomposes the weakly supervised localization task into two complementary heads: a coarse classification head for effective image-level forgery detection and a fine localization head for accurate pixel-level mask prediction. 
The two heads work in a collaborative manner, enabling the model to gradually improve its ability to detect tampered regions.

\subsubsection{Coarse Classification Head.} 
Given the text-enhanced visual features $F^I_{enh}$ obtained from the adaptive vision-language reasoning module, we first predict the tampering probability of each image patch through a binary classifier. 
To focus on the most suspicious regions, we select the top-$K$ patches with the highest tampering probability and aggregate their scores to form an image-level prediction $\hat{y}_{coarse}$. 
The top-$K$ pooling strategy simulates the weakly supervised setting, where only a few localized patches are manipulated while the remaining are authentic.
The coarse classification loss is defined as a binary cross-entropy (BCE) loss:
\begin{equation}
    \mathcal{L}_{coarse} = -\big[y \log(\hat{y}_{coarse}) + (1-y) \log(1-\hat{y}_{coarse})\big],
\end{equation}
where $y \in \{0, 1\}$ is the ground-truth image-level label. This branch ensures that the network can effectively distinguish forged images from pristine ones.

\subsubsection{Fine Localization Head.} 

To achieve pixel-level localization, we design a fine localization head that decodes patch-level features into a pixel-level manipulation mask. 
Specifically, the enhanced visual features $F^I_{enh}$ and the tamper-resistant text embeddings $F^T_{ta}$ are combined via a dot product operation to obtain a cross-modal similarity map, which is then input into a masking decoder to predict pixel-level masks $\hat{M}$.

To leverage weak image-level supervision while learning spatially discriminative features, we introduce an adaptive soft-gated pooling (SG pooling) layer on top of $\hat{M}$. 
Unlike conventional max or average pooling, which either focus excessively on peak responses or dilute discriminative pixel clues, SG pooling employs a differentiable gating mechanism with learnable threshold and temperature parameters. 
The mechanism adaptively assigns higher weights to manipulated pixels and suppresses background responses, enabling robust aggregation of pixel-level predictions into an image-level score $\hat{y}_{fine}$. 
By maintaining differentiability, SG pooling facilitates end-to-end optimization and allows the network to emphasize manipulation-relevant pixels automatically. 
The localization loss for this branch is defined as a binary cross-entropy loss:
\begin{equation}
    \mathcal{L}_{fine} = -\big[y \log(\hat{y}_{fine}) + (1-y) \log(1-\hat{y}_{fine})\big],
\end{equation}
where $y \in \{0, 1\}$ is the image-level ground-truth label. 

By jointly optimizing $\mathcal{L}_{coarse}$ and $\mathcal{L}_{fine}$, the dual-branch architecture enables the model to first coarsely determine whether an image contains manipulations and then further identify pixel-level tampered regions.

\subsection{Contrastive Patch Consistency Constraint}

% Although the dual-branch architecture provides coarse-to-fine supervision, patch-level features can still be inconsistent, particularly under weak supervision where pixel-level annotations are unavailable. 
% To address this, we propose a contrastive patch consistency (CPC) constaint that explicitly enforces spatial coherence among patch embeddings, encouraging patches with similar manipulation clues to cluster together while pushing apart features from authentic regions. This mechanism improves the discriminability of learned image representations and enhances the quality of the predicted tampering mask.

Although the dual-branch architecture provides predictions from coarse to fine, the absence of GT pixel-level labels still hinders the distinction between real and fake parts of image features. 
Therefore, based on the idea of self-supervised learning, we propose a contrastive patch consistency (CPC) constraint, which encourages parts with similar forgery clues to cluster together while pushing away parts from the true region. 
The constraint enhances the discriminative ability of the learned image representations and improves the quality of the predicted forgery mask.

% Given the patch-level enhanced visual features $F^I_{enh}$ and the predicted manipulation mask $\hat{M}$, we assign a pseudo label to each patch based on its average tampering probability. 
Based on the patch-level enhanced visual feature $F^I_{enh}$, we assign a label to each patch based on the predicted manipulation mask $\hat{M}$.
Specifically, patches with responses above a threshold $\tau_{fg}$ are treated as \emph{tampered}, while those below $\tau_{bg}$ are treated as \emph{authentic}. 
This labeling allows us to build positive and negative patch pairs without requiring ground-truth masks.

Concretely, for a tampered patch feature $f_i^{tam}$ and an authentic patch feature $f_j^{real}$, we compute the similarity scores via normalized dot products. The ${L}_{cpc}$is formulated as:
\begin{align}
    \mathcal{L}_{cpc} &= -\frac{1}{|\mathcal{P}|}\sum_{(i,j)\in\mathcal{P}}
\Big[
\log\frac{\exp(\text{sim}(f_i^{tam}, f_j^{tam})/\gamma)}%
{\sum_{k}\exp(\text{sim}(f_i^{tam}, f_k)/\gamma)} \\ \notag
&+
\log\frac{\exp(\text{sim}(f_j^{real}, f_i^{real})/\gamma)}%
{\sum_{k}\exp(\text{sim}(f_j^{real}, f_k)/\gamma)}
\Big],
\end{align}
where $\mathcal{P}$ denotes the set of positive patch pairs, $\gamma$ is a temperature parameter, and $\text{sim}(\cdot,\cdot)$ represents cosine similarity. 
The first term pulls together features from tampered patches, while the second term aligns authentic patches. 
Negatives are sampled from patches of the opposite type, thereby pushing apart dissimilar features.

% Unlike previous weakly supervised approaches that rely solely on image-level losses, CPC directly regularizes spatial representations at the patch level, providing a surrogate supervision signal for mask learning. This design encourages feature clustering for manipulated regions and spatial smoothness across the predicted mask, ultimately leading to more accurate and coherent forgery localization.

\subsection{Objective Function}

The entire framework is trained end-to-end under weak supervision by jointly optimizing the losses from all predicted heads. 
The total loss is defined as:
\begin{equation}
    \mathcal{L} = \mathcal{L}_{coarse} + \mathcal{L}_{fine} + \lambda_{ccs}(t)\,\mathcal{L}_{cpc},
\end{equation}
where $\mathcal{L}_{coarse}$ and $\mathcal{L}_{fine}$ are the binary classification losses from the coarse classification head and fine localization head, respectively. 
$\mathcal{L}_{cpc}$ is the contrastive patch consistency constraint that regularizes patch-level feature learning. 
Unlike a fixed weighting, we adopt a warm-up scheduling strategy \cite{goyal2017accurate} for the consistency term: 
\begin{equation}
\lambda_{ccs}(t) =
\begin{cases}
0, & t < T_w, \\
1 - \exp\big(-\frac{t - T_w}{T_{total} - T_w}\big), & t \geq T_w,
\end{cases}
\end{equation}
where $t$ is the current training epoch, $T_w$ is the warm-up starting epoch, and $T_{total}$ is the total number of training epochs. 
This schedule gradually activates the contrastive patch consistency constraint after an initial warm-up period, allowing the network to first learn stable image-level discrimination before enforcing spatial consistency. 
All modules of ViLaCo are optimized jointly via stochastic gradient descent with backpropagation.

\begin{table*}[]
\centering
\resizebox{0.95\textwidth}{!}{\begin{tabular}{llcccccc|ccccc}
\toprule
& \multirow{2}{*}{\textbf{Baselines}} & \multicolumn{6}{c}{\textbf{Pixel-Level F1}} & \multicolumn{5}{c}{\textbf{Combined F1}}\\
&  &  \textbf{CASIAv1} & \textbf{Columbia} & \textbf{Coverage} & \textbf{IMD2020} & \textbf{NIST16} & \textbf{AVG} & \textbf{CASIAv1} & \textbf{Columbia} & \textbf{Coverage} & \textbf{IMD2020}  & \textbf{AVG} \\
\midrule
\multirow{2}{*}{\rotatebox{90}{\textbf{Un.}}} & NOI1  & 0.157 & 0.311 & 0.205 & 0.124 & 0.089 & 0.190 & 0.000 & 0.000 & 0.000 & 0.000 & 0.000 \\
 & CFA1  & 0.140 & 0.320 & 0.188 & 0.111 & 0.106 & 0.188  & 0.000 & 0.000 & 0.000 & 0.000 & 0.000 \\
\midrule
\multirow{9}{*}{\rotatebox{90}{\textbf{Fully-supervised}}} &H-LSTM&  0.154&  0.130&  0.163&  0.195&  0.354&  0.176&  0.000&  0.004&  0.000&  0.000&  0.001  \\ 
& ManTra-Net&  0.155&   0.364&  0.286&  0.122&  0.000&  0.185&  0.000&  0.000&  0.000&  0.000&  0.000  \\ 
& RRU-Net&   0.225&   0.452&  0.189&  0.232&  0.265&  0.273&  0.023&  0.000&  0.000&  0.000&  0.006  \\ 
& CR-CNN&  0.405&   0.436&  0.291&  -&  0.238&  -&  0.382&  0.413&  0.181&  -&  -  \\ 
& GSR-Net&  0.387&   0.613&  0.285&  0.175&  0.283&  0.349&  0.042&  0.042&  0.000&  0.026&  0.028  \\ 
& SPAN&  0.184&   0.487&  0.172&  0.170&  0.221&  0.214&  0.000&  0.000&  0.000&  0.000&  0.000  \\ 
& CAT-Net&   0.276&   0.352&  0.134&  0.102&  0.138&  0.200&  0.345&  0.406&  0.149&  0.144&  0.261  \\ 
& MVSS-Net&  0452&   0.638&  0.453&  0.260&  0.292&  0.419&  0.566&  0.711&  0.317&  0.300&  0.474 \\ 
& IF-OSN&  0.686 & 0.728  &  0.743 & 0.576 & 0.645 & 0.676  & 0.857 & 0.904 & 0.678 & 0.547 & 0.747   \\ 
\midrule

\multirow{8}{*}{\rotatebox{90}{\textbf{Weakly-supervised}}} & MIL-FCN&  0.117&   0.089&  0.121&  0.097&  0.024&  0.090&  0.193&  0.141&  0.118&  0.131&  0.146  \\  
&  MIL-FCN+WSCL&  0.172&   0.270&  0.178&  0.193&  0.110&  0.185&  0.280&  0.386&  0.268&  0.252&  0.296  \\ 
& Araslanov&   0.112&   0.102&  0.127&  0.094&  0.026&  0.092&  0.194&  0.140&  0.133&  0.046&  0.125  \\ 
& Araslanov+WSCL&  0.153&   0.362&  0.201&  0.173&  0.099&  0.198&  0.250&  0.414&  0.255&  0.159&  0.270  \\ 
& EdgeCAM& 0.338 & 0.470  & 0.262 & 0.242 &0.254 & 0.313 & 0.476 & 0.573  & 0.297 & 0.347 & 0.423   \\ 
& WSCCL& \underline{0.347} & 0.273  & \underline{0.301} & \underline{0.265} & \underline{0.159} & \underline{0.269} & 0.475  & 0.302 & \underline{0.427} & \underline{0.365} & 0.388    \\ 
& MRL-Net&  \underline{0.347} &   \underline{0.534}&  0.213&  0.248&  0.113&  0.265&  \underline{0.495}&  \underline{0.603}&  0.316&  0.348&  \underline{0.441} \\ 
& \cellcolor{light-gray}ViLaCo   & \cellcolor{light-gray}\textbf{0.491} & \cellcolor{light-gray}\textbf{0.536} & \cellcolor{light-gray}\textbf{0.319} & \cellcolor{light-gray}\textbf{0.365} & \cellcolor{light-gray}\textbf{0.267} & \cellcolor{light-gray}\textbf{0.373} & \cellcolor{light-gray}\textbf{0.632} & \cellcolor{light-gray}\textbf{0.714} & \cellcolor{light-gray}\textbf{0.568} & \cellcolor{light-gray}\textbf{0.456} & \cellcolor{light-gray}\textbf{0.593}  \\
\bottomrule
\end{tabular}} 
\caption{Comparison with unsupervised (Un.), fully-supervised and weakly-supervised methods on pixel-level manipulation localization pF1 score  and the combined F1 score between I-F1 and P-F1. The best and the second best results in weakly-supervised methods are noted with \textbf{bolded} and \underline{underlined} respectively.}
\label{tab:Pixel Level comparisons}
\end{table*}

\section{Experiments}

\subsection{Setup}

\subsubsection{Datasets}
% 训练时，保证样本的均衡性

For consistency and fairness, we follow the settings of previous weakly supervised image localization methods \cite{zhai2023towards, li2025m2rl}.
Our experiments are trained on the CASIAv2 dataset \cite{dong2013casia}, with in-dataset testing on CASIAv1 and cross-dataset testing on Columbia \cite{hsu2006detecting}, COVER \cite{wen2016coverage}, NIST16 \cite{guan2019mfc}, and IMD20 \cite{novozamsky2020imd2020}.

\subsubsection{Evaluation Metrics}

We assess localization performance using pixel-level F1 (P-F1) for manipulated regions and combined F1 (C-F1) for overall accuracy, both computed with a fixed threshold of 0.5. Image-level detection is evaluated using image F1 (I-F1).

\subsubsection{Implementation Details.}

Our network adopts frozen image and text encoders from the pre-trained CLIP (ViT-B/16), with transformer-based FFN layers where ReLU activations are replaced by GELU. 
All input images are resized to $256 \times 256$ and augmented via standard cropping and flipping. The patch size is set to $8 \times 8$.
The model is implemented in PyTorch and trained on a single NVIDIA RTX 4090 GPU using the AdamW optimizer \cite{loshchilov2017decoupled} with a batch size of 32, an initial learning rate of 0.0001, and a total of 100 epochs.

\subsection{Comparison with State-of-the-Art}

In this section, we compare ViLaCo’s image-level detection and pixel-level localization performance with 18 existing methods. Unsupervised: CFA1 \cite{ferrara2012image}, NOI1 \cite{mahdian2009using}; fully supervised: H-LSTM \cite{bappy2019hybrid}, ManTra-Net \cite{wu2019mantra}, RRU-Net \cite{bi2019rru}, CR-CNN \cite{yang2019constrained}, GSR-Net \cite{zhou2020generate}, SPAN \cite{hu2020span}, CAT-Net \cite{kwon2022learning}, FCN+DA \cite{chen2021image}, MVSS-Net \cite{dong2022mvss}, IF-OSN \cite{wu2022robust}; weakly supervised: MIL-FCN \cite{pathak2014fully}, Araslanov \cite{araslanov2020single}, WSCL \cite{zhai2023towards}, EdgeCAM \cite{zhou2024exploring}, WSCCL \cite{bai2025weakly}, MRL-Net \cite{li2025m2rl}.

%介绍一下数据集

\subsubsection{Pixel-Level Localization Comparisons}

The pixel-level localization results are presented in Tab.~\ref{tab:Pixel Level comparisons}, VaLiCo is compared with unsupervised, fully supervised, and weakly supervised baselines across five datasets.
ViLaCo achieves state-of-the-art results among weakly-supervised approaches, significantly outperforming counterparts in both pixel-level F1 and combined F1 scores. Moreover, our approach shows strong competitiveness compared to fully-supervised methods; notably, ViLaCo improves over MVSS-Net on average Combined F1 by $11.9\%$. These observations demonstrate that ViLaCo not only achieves accurate pixel-level localization under limited supervision but also exhibits superior generalization and robustness to diverse manipulation scenarios.

% 可以不用加上detection
\subsubsection{Image-Level Detection Comparisons}

% To further evaluate the effectiveness of our method in the image-level detection task, we also compare ViLaCo against state-of-the-art baselines in Tab.~\ref{tab:ImageLevel comparisons}. ViLaCo consistently outperforms other weakly-supervised methods, achieving the best average Image-level F1 (I-F1) of 0.776. Specifically, our method surpasses the second-best weakly-supervised approach, MRL-Net, by $1.4\%$ on CASIAv1 and by $9.0\%$ on IMD20. Moreover, ViLaCo demonstrates robust performance competitive to fully-supervised methods, exceeding MVSS-Net's performance by an average improvement of $24.2\%$. This further validates the superiority of ViLaCo in accurately distinguishing manipulated from authentic images, even under limited supervision.

To assess image-level detection performance, we compare ViLaCo with state-of-the-art baselines (Tab.~\ref{tab:ImageLevel comparisons}). ViLaCo achieves the highest average I-F1 score of 0.776, consistently outperforming weakly supervised methods and remaining competitive with fully supervised approaches, further validating its effectiveness under limited supervision.

\subsection{Qualitative Results}

\begin{figure}
\centering
\includegraphics[width=0.4\textwidth]{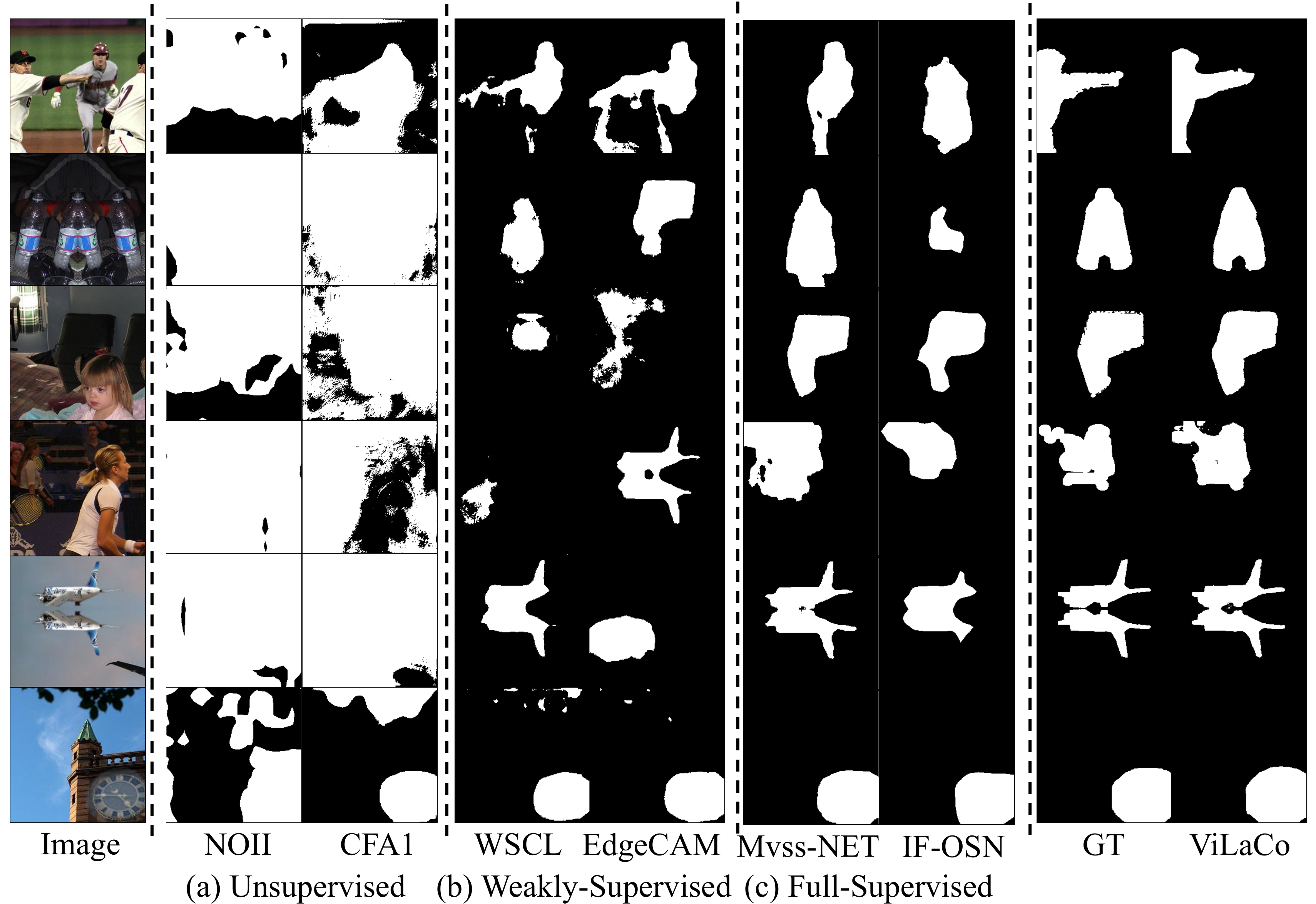}
\caption{Qualitative comparison of ViLaCo on DEFACTO datasets with (a) unsupervised, (b) weakly-supervised and (c) fully-supervised methods.}
\label{fig:loc_visual}
\end{figure}

We further provide qualitative results to visually illustrate the effectiveness of ViLaCo in localizing manipulated regions. As depicted in Fig. \ref{fig:loc_visual}, our method demonstrates superior performance compared to both unsupervised and existing weakly-supervised methods. 
% Specifically, the unsupervised approaches (NOI1, CFA1) generate fragmented and noisy predictions. Meanwhile, weakly-supervised methods like WSCL and EdgeCAM predict excessively large regions, which partially cover the ground truth but lack precision. Fully-supervised methods (e.g., MVSS-Net, IF-OSN) provide clearer localization maps; however, ViLaCo achieves comparable or even superior localization quality despite only weak supervision, closely matching the ground truth masks. 
Compared with existing weakly-supervised methods, ViLaCo achieves significantly better localization performance and reaches a quality comparable to fully-supervised methods.
This result highlights ViLaCo’s capability to produce clearer and more accurate pixel-level localization masks by leveraging vision-language alignment.

\begin{figure}
    \centering
    \subfigure[block size.]{%
        \includegraphics[width=0.22\textwidth]{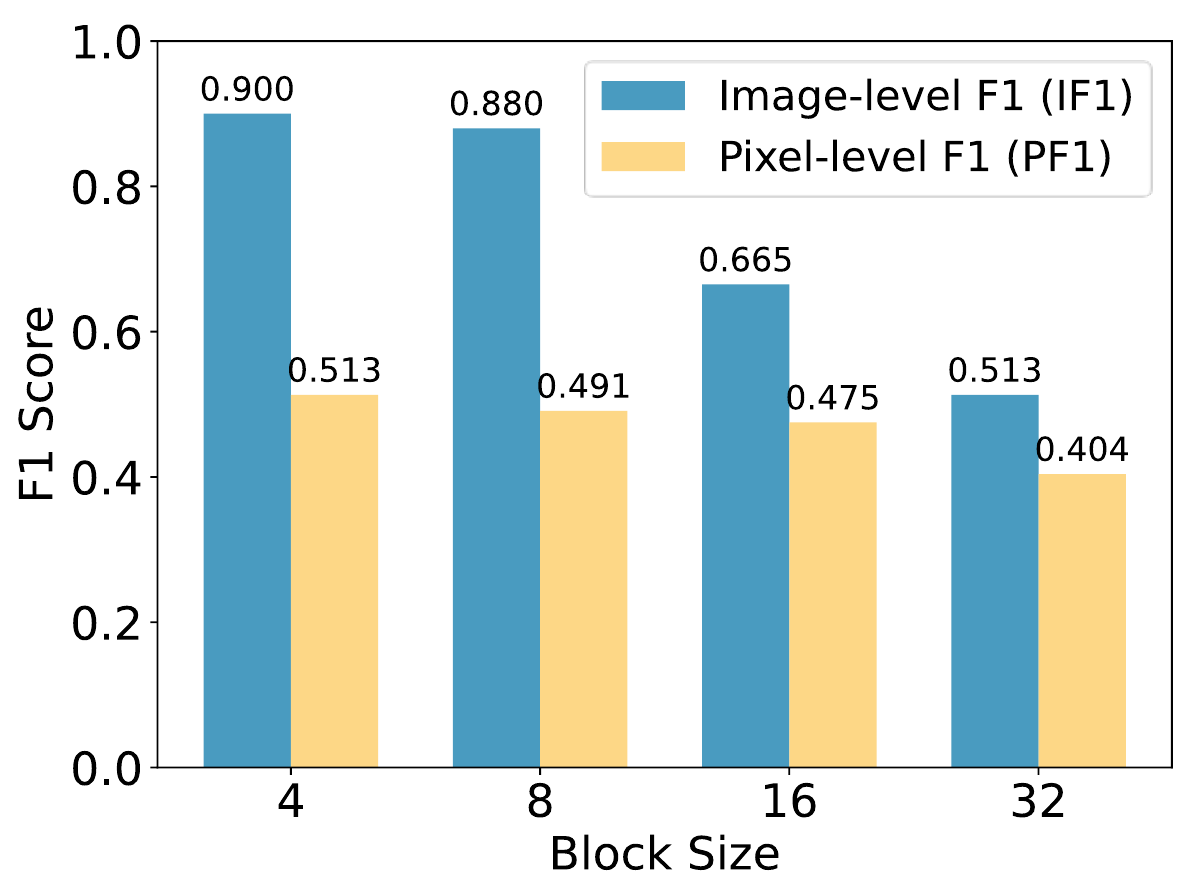}%
        \label{fig:blocksize_F1}
    }
    \subfigure[ Two prediction heads.]{%
        \includegraphics[width=0.22\textwidth]{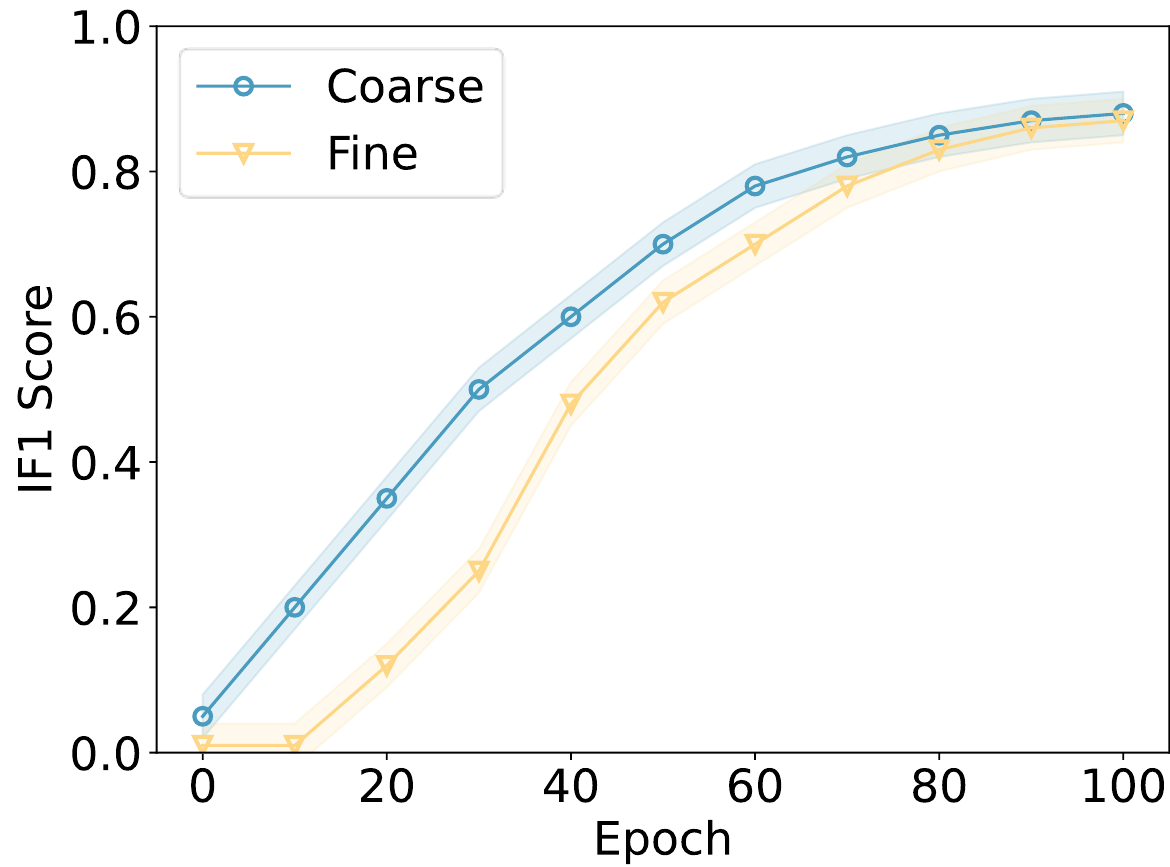}%
        \label{fig:branch_choose}
    }
    \caption{Effect of block size and the prediction head on detection and localization performance}
    \label{fig:visual_two}
\end{figure}

In addition, we analyze the impact of block size and the two prediction heads on detection and localization performance, as shown in Fig.~\ref{fig:visual_two}.
We observe that selecting a block size of $8 \times 8$ offers the best trade-off between localization accuracy and computational efficiency. Moreover, as training progresses, the performance of the coarse and fine prediction heads gradually converges, indicating that both branches collaboratively contribute to improved and stable detection performance.

\subsection{Ablation Study}

\begin{table}
\centering
\resizebox{0.45\textwidth}{!}{\begin{tabular}{l l c c c c c}
\toprule
& \textbf{Baselines}   &\textbf{CASIAv1} & \textbf{Columbia}  & \textbf{COVER} & \textbf{IMD20}  & \textbf{AVG} \\
\midrule
% \textit{\textbf{Unsupervised}} & &   &  &  &  \\
\multirow{2}{*}{\rotatebox{90}{\textbf{Un.}}} &
NOI1   & 0.000  & 0.000  & 0.000  & 0.000  & 0.000  \\ 
& CFA1   & 0.000  & 0.000  & 0.000 & 0.000  & 0.000\\
\midrule
\multirow{9}{*}{\rotatebox{90}{\textbf{Fully.}}} 
& H-LSTM                     & 0.000  & 0.002  & 0.000  & 0.000  & 0.001 \\ 
& ManTra-Net                  & 0.000  & 0.000  & 0.000  & 0.000  & 0.000  \\ 
& RRU-Net                     & 0.001  & 0.000  & 0.000  & 0.000  & 0.000  \\ 
& CR-CNN                     & 0.361  & 0.392  & 0.131  & 0.200 & 0.271  \\ 
& GSR-Net                      & 0.022  & 0.022  & 0.000  & 0.014 & 0.0019 \\ 
& SPAN                         & 0.000  & 0.000  & 0.000  & 0.000  & 0.000 \\ 
& CAT-Net                      & 0.459  & 0.505  & 0.169  & 0.229  & 0.157 \\ 
& FCN+DA                       & 0.775  & 0.481  & 0.180  & 0.182  & 0.404 \\ 
& MVSS-Net                    & 0.758  & 0.802  & 0.244  & 0.355  & 0.534 \\ 
\midrule
\multirow{6}{*}{\rotatebox{90}{\textbf{Weakly.}}}
& MIL-FCN                      & 0.553  & 0.338  & 0.115  & 0.205  & 0.303  \\ 
& MIL-FCN+WSCL                 & 0.738  & 0.680  & 0.544  & 0.360  & 0.580  \\ 
& Araslanov                    & 0.496  & 0.140  & 0.140  & 0.219  & 0.270  \\ 
& Araslanov+WSCL               & 0.679  & 0.483  & 0.348 & 0.316  & 0.456  \\ 
& EdgeCAM                      & 0.806  & 0.733  & 0.343 & \underline{0.613}  & 0.624 \\ 
& MRL-Net                     & \underline{0.866}  & \textbf{0.975}  & \underline{0.610}  & 0.585  & \underline{0.762}  \\ 
& \cellcolor{light-gray}ViLaCo   & \cellcolor{light-gray}\textbf{0.880 } & \cellcolor{light-gray}\underline{0.917} & \cellcolor{light-gray}\textbf{0.630} & \cellcolor{light-gray}\textbf{0.675} & \cellcolor{light-gray}\textbf{0.776} \\
\bottomrule
\end{tabular}} 
\caption{Comparison of state-of-the-art methods for image-level manipulation detection across multiple datasets, evaluated by image-level F1 score. The first and second rankings are shown in \textbf{bolded} and \underline{underlined} respectively.}
\label{tab:ImageLevel comparisons}
\end{table}

\begin{table}
\centering
\resizebox{0.4\textwidth}{!}{
\begin{tabular}{ccccccccc}
\hline
% Method &F1 &F1\_best &ACC &AUC\\
 \multirow{2}{*}{ID} &\multicolumn{3}{c}{LOSS} &\multicolumn{2}{c}{CASIA} &\multicolumn{2}{c}{IMD20}\\
\cmidrule(lr){2-4}\cmidrule(lr){5-6}\cmidrule(lr){7-8}
 &$\mathcal{L}_{coarse}$  &$\mathcal{L}_{fine}$ &$\mathcal{L}_{cpc}$ &PF1 &IF1  &PF1 &IF1\\
\hline
1&  -& \ding{52}& \ding{52}&  0.842 & 0.410 & 0.620 & 0.305\\
2&  \ding{52}&  -& \ding{52}&  0.790 & 0.365 & 0.575 & 0.270\\
3&  \ding{52}& \ding{52}& -&  0.860 & 0.445 & 0.640 & 0.320\\
4&  \ding{52}& \ding{52}& \ding{52}& \textbf{0.880} & \textbf{0.491 }&\textbf{0.675} &\textbf{0.365}\\ 
\hline
\end{tabular}
}
\caption{Abalation study of the proposed $\mathcal{L}_{coarse}$, $\mathcal{L}_{fine}$ and and $\mathcal{L}_{cpc}$ in terms of F1 score. The bold mark best performance}
\label{tab: ablation study}
\end{table}

To investigate the contribution of each loss function in ViLaCo, we perform an ablation study by selectively removing $\mathcal{L}_{coarse}$, $\mathcal{L}_{fine}$, and $\mathcal{L}_{cpc}$, and report the results in Tab.~\ref{tab: ablation study}.

When $\mathcal{L}_{coarse}$ is removed (ID 1), the model relies solely on fine-branch supervision, leading to a noticeable drop in image-level detection accuracy (IF1 decreases from 0.491 to 0.410 on CASIA). 
Meanwhile, removing $\mathcal{L}_{fine}$ (ID 2) causes a similar degradation, indicating that both branches play complementary roles in improving localization precision.
Excluding the contrastive consistency loss $\mathcal{L}_{cpc}$ (ID 3) also reduces performance, particularly on IMD20, where IF1 drops from 0.365 to 0.320, showing that $\mathcal{L}_{cpc}$ effectively enhances the model’s ability to distinguish manipulated regions from authentic ones.
Finally, when all three losses are combined (ID 4), ViLaCo achieves the best results across both datasets and metrics, demonstrating that the joint optimization of these losses is crucial for achieving accurate pixel-level and image-level localization.

\section{Conclusion}

In this work, we presented ViLaCo, the vision-language collaborative reasoning framework for weakly supervised image forgery localization. Unlike prior WSIFL approaches that solely rely on intra-image consistency clues, ViLaCo introduces auxiliary semantic supervision distilled from pre-trained vision-language models, enabling precise localization using only image-level annotations. The proposed architecture progressively integrates this semantic knowledge through three key designs: a vision-language feature modeling network with a local-global spatial adapter to capture forgery-specific clues, an adaptive vision-language reasoning network that fuses textual semantics and visual features to highlight manipulated regions, and a coarse-to-fine dual prediction mechanism enhanced by a contrastive patch consistency loss to refine mask quality.
Extensive experiments across multiple benchmarks demonstrate that ViLaCo significantly outperforms existing weakly supervised methods and achieves competitive performance compared to fully supervised counterparts. 

\bibliography{aaai2026}

\end{document}